# Computing the Hazard Ratios Associated with Explanatory Variables Using Machine Learning Models of Survival Data

**Authors:** Sameer Sundrani[1,2], James Lu[1*]

**Affiliations:**

[1]Modeling & Simulation/Clinical Pharmacology, Genentech, South San Francisco, California

[2]Biomedical Computation, Schools of Engineering and Medicine, Stanford University, Stanford, California

**\*Corresponding author**:
James Lu
Genentech, 1 DNA Way,
South San Francisco, CA, 94080
Email: lu.james@gene.com
Phone: 650-6199325

**ABSTRACT**  (275 words max)

**Purpose:** The application of Cox Proportional Hazards (CoxPH) models to survival data and the derivation of Hazard Ratio (HR) is well established. While nonlinear, tree-based Machine Learning (ML) models have been developed and applied to the survival analysis, no methodology exists for computing HRs associated with explanatory variables from such models. We describe a novel way to compute HRs from tree-based ML models using the Shapley additive explanation (SHAP) values, which is a locally accurate and consistent methodology to quantify explanatory variables' contribution to predictions.

**Methods:** We used three sets of publicly available survival data consisting of patients with colon, breast or pan cancer and compared the performance of CoxPH to the state-of-art ML model, XGBoost. To compute the HR for explanatory variables from the XGBoost model, the SHAP values were exponentiated and the ratio of the means over the two subgroups calculated. The confidence interval was computed via bootstrapping the training data and generating the ML model 1000 times. Across the three data sets, we systematically compared HRs for all explanatory variables. Open-source libraries in Python and R were used in the analyses.

**Results:** For the colon and breast cancer data sets, the performance of CoxPH and XGBoost were comparable and we showed good consistency in the computed HRs. In the pan-cancer dataset, we showed agreement in most variables but also an opposite finding in two of the explanatory variables between the CoxPH and XGBoost result. Subsequent Kaplan-Meier plots supported the finding of the XGBoost model.

**Conclusion:** Enabling the derivation of HR from ML models can help to improve the identification of risk factors from complex survival datasets and enhance the prediction of clinical trial outcomes.

# INTRODUCTION

The use of Cox Proportional Hazards (CoxPH) model to survival data is well established.[1,2] In particular, the hazard function for a patient *i* with the vector of explanatory variables $x_i = (x_{1i}, x_{2i}, \ldots, x_{pi})$ can be expressed as:

$$h_i(t) = exp(\beta_1 x_{1i} + \beta_2 x_{2i} + \ldots + \beta_p x_{pi}) \times h_0(t), \tag{1}$$

where $h_0(t)$ is the baseline hazard function and $\beta$ the vector of coefficients of explanatory variables. Once the coefficient vector $\beta$ is estimated from the survival data, it can be interpreted as the logarithms of the ratio of the hazard of death to the baseline hazard $h_0(t)$.[1] That is, the hazard ratios (HRs) and the corresponding confidence intervals with respect to the explanatory variables can be directly obtained from the coefficient vector of the fitted model. The ease of interpretation of the CoxPH model results underlies its popularity and broad acceptance in the medical community.

Recent years have witnessed a significant increase in the use of machine learning (ML) in the healthcare setting[3] and this interest has spurred the development of a wide variety of ML methodologies for survival data.[4] Within the field of oncology, several research efforts have demonstrated how ML can complement statistical and mechanistic modeling analysis of data.[5-8] While ML has the potential to transform oncology[9], there are a number of obstacles to its broader adoption. One key challenge is that of gaining the trust of clinicians through its interpretability. In particular, while nonlinear ensemble methods such as random forests[10] and XGBoost[11] have the benefit of improved predictions through aggregation of individual trees that can capture more complex interactions, they also have the drawback of being less interpretable as a result of those complex aggregations.[9] However, recent advances in explainability have helped to elucidate these models and extract insights.[9] Prominent amongst those is the development of SHapley Additive exPlanations (SHAP), which provides optimal explanations of nonlinear tree-based ML models based on game theory.[12] In particular, SHAP values demonstrate *local accuracy* and *consistency*.[12] In the medical context, the use of SHAP values have been demonstrated with ML analysis of clinical data from the National Health and Nutrition Examination Survey (NHANES) I Epidemiologic Follow-up Study[13] as well as from the Chronic Renal Insufficiency Cohort (CRIC) study.[12] Within the field of oncology, SHAP analysis has found application in the analysis of overall survival (OS) in 372,808 prostate cancer patients by helping to interpret the predictions and visualize interactions between explanatory variables of XGBoost, a tree-based ML model.[6] The nonlinear relationships inferred from SHAP analysis were subsequently confirmed using Kaplan-Meier estimates and CoxPH[6], and supported by prior literature.

Despite the advances offered by SHAP analysis to explain predictions of nonlinear ML models, as has been noted[6] there is currently no established way to interpret statistical significance of explanatory variables from ML models in a manner comparable to p-values from CoxPH models, pointing to a need for further development. Indeed, this current limitation of reporting the significance of explanatory variables inferred from ML models prevented a direct comparison of CoxPH against ML models.[8] In this work, we propose a way to compute from ML models of patient survival data the HR and the confidence intervals (CI) for how explanatory variables affect OS. We demonstrate that the built-in properties of SHAP values[12] enable, not only an interpretation of how ML models utilize the explanatory variables, but also a quantitation of the explanatory variables' impact on the hazard function. By bootstrapping the ML via resampling the training data, one can additionally derive the CI of the HR. We demonstrate the proposed methodology on three publicly available patient survival datasets for the following respective cancer types: breast[14], colon[15], and pan-cancer.[16]

**METHODS**

*Simulated Dataset*

As a consistency check of our proposed methodology, we simulated a survival dataset of 850 synthetic subjects (an arbitrary number to produce a sizeable dataset) with three binary explanatory variables ('var1', 'var2', and 'var3') where the linear coefficients were set to $\beta_{var1} = 1$, $\beta_{var2} = -2$, and $\beta_{var3} = 2$. In this dataset, 20% of subjects were right-censored and simulated patients were allowed to take survival times ranging over 10,000 days. We created this dataset utilizing the R 'sim.survdata' package (see code contained in Supplementary Data[32]).[17]

*Publicly available Datasets*

In this work, we've applied the proposed methodology on the three data sets detailed as follows. The breast cancer survival dataset[14] is drawn from a prospective study performed by the German Breast Cancer Study Group analyzing traditional prognostic factors in patients with node-positive breast cancer. The authors originally applied a fractional polynomial model to predict the clinical progression of studied patients. The colon cancer[15] dataset describes survival results from one of the first successful adjuvant chemotherapy trials for colon cancer. In this study, patients with stage B and C colorectal carcinoma were randomly assigned to three treatment groups: no further treatment, treatment with levamisole alone, or treatment with levamisole and fluorouracil. While this dataset included both recurrence and death records for each patient, we only examined those records for death. Finally, the pan-cancer[16] data is drawn from a recent analysis of sequencing data from The Cancer Genome Atlas (TCGA) to further new diagnostics and individualized cancer treatments. The authors reported findings of 127 significantly mutated genes (SMGs) from a systematic analysis of 3281 tumors across 12 cancer types and examined correlations of clinical features with somatic events across these 127 SMGs within multiple tumor types. In all datasets, we used all available valid patient survival data for ML.

***Derivation of HR from SHAP values***

In the ML approach for Cox survival modeling, the hazard function for a patient *i* with the vector of explanatory variables $x_i = (x_{1i}, x_{2i}, \ldots, x_{pi})$ can be expressed as:

$$h_i(t) = exp(f(x_i)) \times h_0(t), \tag{2}$$

where the function $f$ is approximated by the ML model. For instance, in the case of XGBoost, $f$ is approximated by an ensemble of trees, which is typically complex to interpret.

Through SHAP analysis [Lundberg 2020], the function $f$ is decomposed into the following manner:

$$f(x_i) = \phi_0 + \sum_{j=1}^{p} \phi_j(f, x_i), \qquad (3)$$

where $\phi_j(f, x_i)$ is the SHAP-value of explanatory variable $j$ for patient $i$. Via eqn. (2), we can then express the hazard function as:

$$h_i(t) = exp(\phi_1(f, x_i)) \times exp(\phi_2(f, x_i)) \times \ldots \times exp(\phi_0) \times h_0(t), \qquad (4)$$

which shows that the HR associated with variable $j$ can be computed from averaging $exp(\phi_j(f, x_i))$ over patients $i$ within subgroups (for instance, patients who are either positive or negative for a genetic mutation) in a manner analogous to CoxPH.

*Data Preparation and Modeling*

All R and Python code used to generate results and figures are available in the Supplementary Data[32] (see the code contained therein). Initial dataset preparation and CoxPH modeling was done in R (Version 4.0.2) utilizing the 'survival' package.[18]

Subsequent XGBoost modeling was performed in Python (Version 3.7) using the 'xgboost' and 'shap' packages and evaluated in part with the 'scikit-learn' library.[11,19]

On each of the datasets, string-based categorical variables were transformed into numerical values, and each continuous variable was normalized by the range of its respective values to the range [0,1]. To run survival prediction in XGBoost, the survival time in the original dataset, $T_i$, for each patient $i$ was transformed to $T'_i$ according to their censoring information where $C_i = 1$ if patient $i$ was not censored and $C_i = -1$ otherwise.

$$T'_i = C_i \times T_i. \tag{5}$$

For the breast cancer dataset, we did not include the date of diagnosis ('diagdate') or date of recurrence ('recdate') variables as these were dates, and for the Colon Cancer dataset we did not include the 'study' variable as this was a constant. In the pan-cancer dataset, we included all covariates except for 'Tumor.grade' (due to most of the values being missing), and performed one-hot encoding for each of the tumor types.[20] Patients that did not have appropriate survival information (such as missing survival or censoring dates) were excluded from the data. We also tested three variations of the pan-cancer dataset: (1) including all covariates without imputations; (2) including all covariates with imputations; (3) not including any gene mutation information. For the simulated linear survival dataset, we estimated all three binary variables: 'var1', 'var2', and 'var3' with preset coefficients as mentioned above.

All XGBoost models were trained using the `survival:cox` objective function. Additionally, for each dataset, hyperparameter tuning was performed using the HyperOpt package[22] with 100 evaluation rounds in accordance with parameter ranges that have been previously used in the literature[21], with the exception of capping the `min_child_weight` (minimum sum of instance weight), `reg_alpha` (L1 regularization term on weights) and `reg_lambda` (L2 regularization term on weights) at 10 rather 100. As the objective function for hyperparameter tuning, we computed the mean Harrell's concordance index (C-index) [23] using the 5-fold cross validation approach. Namely, in each evaluation process the total dataset is randomly divided into 5 portions, with the aim of using 4 out of the 5 portions as the training set in the ML model to predict the remaining validation portion and the C-index is computed on the validation portion. For each split of the data into 5-folds, the mean of the set of five C-index values is computed, which serves as the objective value optimized by HyperOpt. Once the hyperparameters have been identified, the model was subsequently evaluated on the same total data set using 5-fold cross validation as explained above, but with the randomly selected folds being distinct from those used in the hyperparameter tuning process in order to accurately assess the generalizability of the model. The full implementation as well as data is provided in the subfolder "XGB_Code_Data" of the Supplementary Data[32].

***Computing the Hazard Ratio and Confidence Intervals***

For ML-derived HR for explanatory variable by two predefined disjoint subgroups (namely, greater than or equal to the median versus below for non-binary variables or 1 versus 0 for binary variables), we took the ratio of the means of the exponentiated SHAP values for the two disjoint subgroups ($S_1, S_2$) as follows:

$$HR_j^{ML} = mean_{i \in S_1}(exp\,(\phi_j(f, x_i)))/\,mean_{i \in S_2}(exp\,(\phi_j(f, x_i))), \qquad (6)$$

where $HR_j^{ML}$ is the ML-derived HR for explanatory variable $j$ and $S_1$ is the first subgroup of interest and $S_2$ is the second (reference) subgroup. For an illustration of the methodology, we show an example with the colon cancer data of SHAP values for both nonbinary and binary explanatory variables where subgroups $S_1$ and $S_2$ are defined by the cut-off given by the variable value (Figure 1). To estimate a 95% confidence interval on each $HR_j^{ML}$ calculated, we performed 1000 bootstraps of this calculation. In each run, we randomly sampled with replacement from the patients to split our dataset into training and test sets, seeding the randomizer with a new value on each run. We then trained a new XGBoost model with the pre-tuned hyperparameters and generated SHAP values for the full dataset. From these SHAP values, we then calculated a HR for each of the explanatory variables following eqn. (6). Finally, we sorted each list and chose the 2.5th and 97.5th percentile values of HR values, respectively.

For each of the processed datasets (as described in the section "Data Preparation and Modeling"), with the exception of the pan-cancer data without imputations, we ran a CoxPH regression as well as 5-Fold cross-validation (CV)[19] in the same manner as for the XGBoost model, and computed the C-index.[23] Namely, in each case the total dataset is divided into 5 portions, whereby each of the 4 portions is used to fit the CoxPH model and its prediction evaluated on the remaining validation portion. For each explanatory variable $j$ in the CoxPH model, we take the estimated coefficient, $\beta_j$, as outputted by the model. To derive the HRs, we take the disjoint subgroups $S_1$ and $S_2$ as defined above for binary and nonbinary explanatory variables and compute the following:

$$HR_j^{CoxPH} = exp\,(\,\beta_j \times [mean_{i\,\in\,S_1}(x_i) \;-\; mean_{i\,\in\,S_2}(x_i)]) \tag{7}$$

where the 95% CI on $HR_j^{CoxPH}$ was generated using the above formula from the standard CoxPH model outputted lower and upper bound $\beta_j$. Note that while for binary variables (where $x_i$ is equal to either 0 or 1) the formula (7) is equivalent to the standard definition of CoxPH-based HR, for non-binary variables the expression is our proposed approach to generalize the concept and derive a HR for the two subgroups.

**RESULTS**

***Proposed HR computation is consistent with CoxPH using simulated data***

On the simulated dataset with preset linear coefficients and only binary variables, we show that XGBoost derived HR computations match all CoxPH results in both direction and significance, with nearly identical median HR approximations and 95% confidence interval bounds (Supplementary Figure 1). C-index comparisons between these two models indicate higher performance for the linear CoxPH model (Table 1 and Supplementary Figure 2).

***Comparison of Model Performance using C-index on Publicly Available Data***

For the breast cancer and colon cancer dataset, we found that XGBoost performed comparably to the CoxPH model when evaluating the distribution of C-indices from the results of a 5-fold CV[20] on unseen data. For the larger pan-cancer dataset, XGBoost demonstrated a trend for higher mean performance, although that difference was not statistically significant (Table 1 and Figure 2).

***Explaining ML Model using SHAP Analysis***

Ranked SHAP value results for the breast cancer data show that the number of progesterone receptors ("prog_recp"), the number of nodes involved ("nodes"), and the number of estrogen receptors ("estrg_recp") were the top 3 features impacting XGBoost predictions (Figure 3a). However, tumor grade ("grade") and menopause status ("menopause") had very little impact on the model outputs. SHAP value rankings by variable for cancer results show that the number of nodes involved ("nodes"), treatment category ("rx"), and patient age were the 3 most important explanatory features while time from surgery to registration ("surg") and colon perforation value ("perfor") had the lowest SHAP-impact (Figure 3b). Similar SHAP rankings for the pan-cancer data without imputations suggest that tumor stage ("Tumor.stage"), age ("Years.to.birth"), and initial tumor diagnosis date ("Date.of.initial.pathological.diagnosis") were the 3 most important explanatory variables with other high ranking features SHAP value showing clear directionality trends for either positive or negative impact on the model output, e.g., positive BRCA tumor type indications ("TCGA.tumor.type.BRCA") show overwhelmingly negative SHAP values (Figure 3c).

***Comparison of HRs***

HR estimations obtained for XGBoost for both the breast cancer and colon cancer datasets were highly consistent with CoxPH results (Figure 4 panels a/b and Supplementary Figure 3). In the breast cancer comparison, the derived HRs agreed in significance at p=0.05 for 7 out of 8 explanatory variables and had overlapping CI for a single variable that did not agree. Of the variables where one or both of the models predicted statistical significance, XGBoost HR predictions agreed with that of CoxPH on 3 out of the 4 explanatory variables and the direction of the median HR effect on outcome was the same for all 4 significant variables. Similarly, in the Colon Cancer comparison, the derived HRs agreed in significance for 8 out of the 10 explanatory variables, and also had overlapping confidence intervals for both variables that disagreed. Of the variables where either one or both of the models predicted significance, XGBoost HR predictions agreed on 4 out of 6 variables for significance and the direction of median HR effect on outcome was identical for all 6 significant variables.

For the pan-cancer data, HR estimates from XGBoost were notably similar to that of CoxPH model in testing all of the following 3 dataset variants: (1) excluding genetic mutation explanatory variables; (2) including those mutation data and imputing median values for missing data; (3) including the mutation data but not imputing any values (Table 1). We found that with our tuned hyperparameters, XGBoost results on the complete data without imputations outperformed both XGBoost and CoxPH results on all the same data sets but with imputation. Our CoxPH result using only a select number of explanatory variables had a slightly higher but not statistically significant mean C-index obtained from 5-fold CV (see Supplementary Figure 4). The CoxPH model also seemed to quickly degrade in predictive power with the addition of all genetic mutation information.

To capture the effects of all 143 explanatory variables, we therefore chose to compare the XGBoost model without imputations to the CoxPH model with imputatons (Figure 4c). There was high similarity between outputs for both models, with agreement on HR direction (median greater than or equal to or less than 1 in 104/143 variables (see Supplementary Figures 5-7). Within the set of explanatory variables where either XGBoost without imputations or CoxPH with imputations found the variable to be significant, the models agreed on median direction in 25/29 variables and both median direction as well as significance at p=0.05 in 11/29. For 2 variables, Tumor Type LUAD and Tumor Type LUSC (representing lung adenocarcinoma and lung squamous cell carcinoma, respectively), XGBoost predicted a direction of variable effect that was in opposition to the CoxPH result. We therefore examined their individual Kaplan-Meier survival curves, which showed that there is a rapid drop in LUAD and LUSC positive patients' survival and the numbers of LUAD and LUSC positive patients are small compared to the LUAD and LUSC negative groups, which is consistent with XGBoost HR estimations (Figure 5 and Supplementary Figure 8).

**DISCUSSION**

While there has been much interest in applying ML to healthcare and drug development applications, several obstacles remain on the path towards broader adoption.[24] While ML methodologies offer some advantages beyond traditional biostatistical methods including flexibility and scalability[25], a framework for establishing the statistical significance of complex ML models remains to be developed.[6] In this work, we show that the same SHAP analysis framework[12] that has been developed to explain ML models[6] can also be used to establish the statistical significance of explanatory variables. In particular, in the context of survival analysis we have developed a way to compute HRs and their CI with respect to explanatory variables, which is part of the output familiar to users of CoxPH analysis. Using our proposed methodology, one can employ ML on survival data and attribute not only the magnitude but also the significance to each of the explanatory variables. As a demonstration of the proposed methodology, we show in the case of simulated survival data our ML-derived HR computation matches that of linear CoxPH results. We further applied the methodology on three sets of clinical survival data and showed good agreement in two of the data sets that contain a small number of explanatory variables, as well as some differences in the large pan-cancer dataset. Amongst the variables that showed discrepancies between the two model results, we examined them further with the results supporting the ML findings. While we demonstrated the use of SHAP analysis for the derivation of HRs only in the context of XGBoost[11] models for survival, note that it can be similarly applied to other types of tree-based ML (such as Random Survival Forest[26]) models as well as neural networks.[27] Demonstration of the proposed methodology to these other ML model types remain an area for further work. Additionally, while we have demonstrated the methodology using the median value as a cut-off to derive the two disjoint subgroups ($S_1, S_2$), the approach can similarly be applied with any selection of the two disjoint subgroups. Some limitations of the proposed SHAP-based HR estimation approach include the potentially large computational efforts required for more complex models, and possible situations where the ML model and the subsequent SHAP analysis do not adequately describe the data. For datasets where the explanatory variables are very high dimensional and little nonlinear effects appear, algorithms such as elastic net for CoxPH may provide accurate and efficient alternatives.[28]

In conclusion, by using the advantages offered by modern ML techniques and delivering the results in a manner familiar to users of biostatistical analyses of survival data, we believe the proposed methodology represents a significant advancement which will help to broaden the adoption of ML and increase its impact within the field of clinical oncology.

The potential applications of the methodology include the identification of risk stratification factors from cancer patients to enable personalized therapy[29,30] and improve the prediction of clinical trial outcomes from tumor growth inhibition metrics.[31]

**ACKNOWLEDGEMENTS**

We would like to acknowledge Dale Miles, Rene Bruno, Kenta Yoshida, Jin Y. Jin, Chunze Li and Amita Joshi for their feedback and support to help make this work possible, as well as Mausumi Debnath, Harbeen Grewal and Srilatha Swami of Anshin Biosolutions for providing editorial assistance.

**Funding information**

This study was performed while the authors were employed by Genentech, Inc.

**TABLES & FIGURES**

Table 1: Summary of the simulated and publicly available datasets used in this study, including variations on the data used and C-index comparison between traditional CoxPH and XGBoost models.

| Dataset | Patients | Events | Number of Explanatory Variables | CoxPH C-Index (Mean± Std.Dev) | XGBoost C-Index (Mean± Std. Dev) |
|---|---|---|---|---|---|
| Simulated Data | 850 | 672 | 3 | 0.790 +/- 0.019 | 0.729 +/- 0.019 |
| Breast Cancer | 686 | 171 | 8 | 0.737 ± 0.027 | 0.747 ± 0.052 |
| Colon Cancer | 888 | 430 | 10 | 0.647 ± 0.031 | 0.663 ± 0.028 |
| Pan-Cancer (limited covs) | 2912 | 984 | 16 | 0.781 ± 0.013 | 0.77 ± 0.018 |
| Pan-Cancer (all, imputing) | 2912 | 984 | 143 | 0.757 ± 0.022 | 0.771 ± 0.019 |
| Pan-Cancer (all, no imputing) | 2912 | 984 | 143 | - | 0.774 ± 0.016 |

**Figure 1**: Illustration of SHAP Dependence and Force Plot Visualization using explanatory variables 'Age' (top) and 'Treatment' (bottom) from the Colon Cancer Data. Disjoint subgroups are divided in the SHAP Dependence Plots by the median value of the variable. Individual force plots show direction of SHAP value effect compared to baseline.

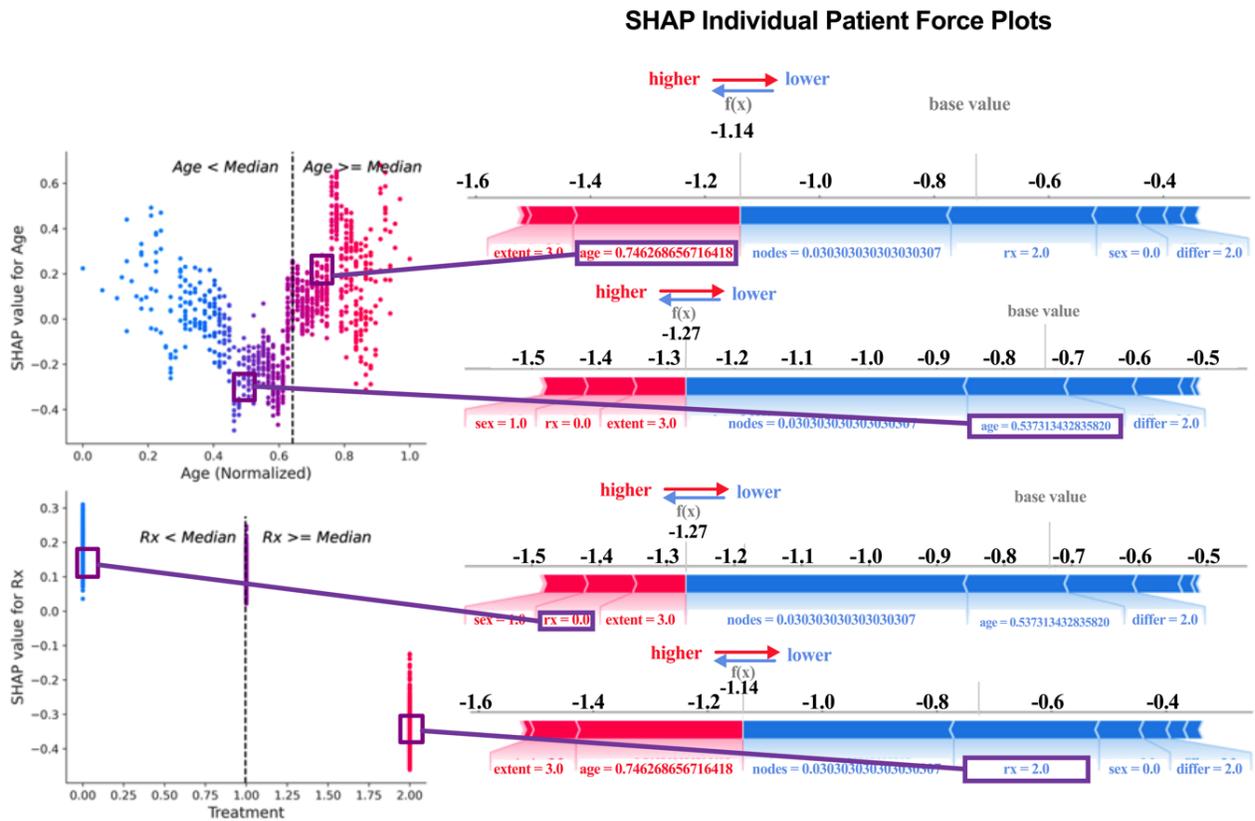

**Figure 2:** C-index comparisons between traditional CoxPH and XGBoost from 5-fold cross validation on each of the datasets. For the pan-cancer data, we evaluated the model performance with no patient left out by comparing CoxPH (with imputation of explanatory variables with missing values) against the XGBoost model without any imputation.

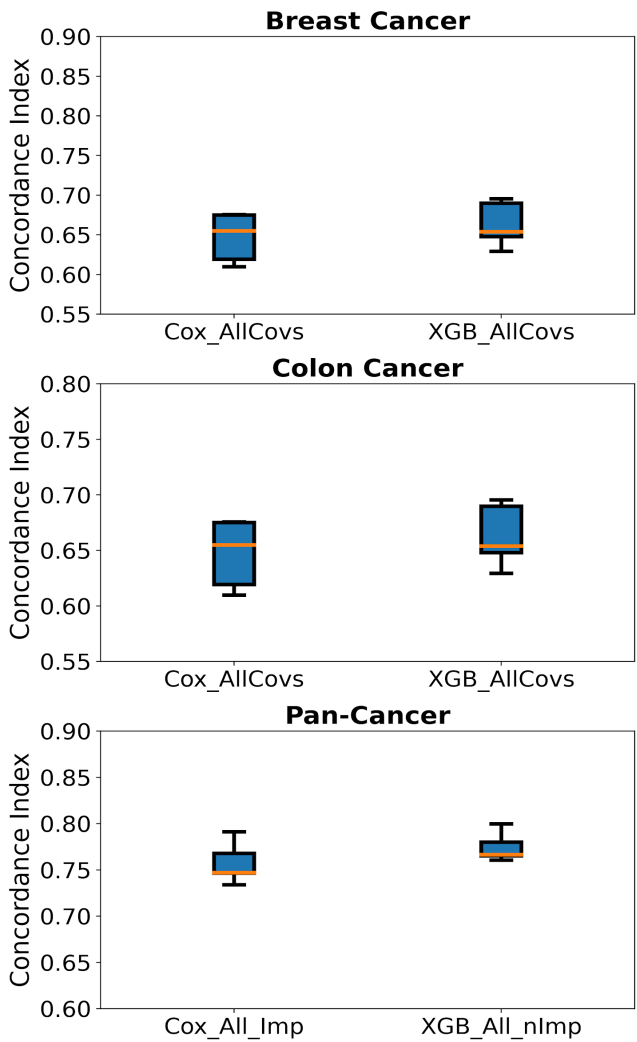

**Figure 3:** Feature importance ranked by the mean absolute magnitude of SHAP values for each of the breast cancer (a), colon cancer (b), and pan-cancer (c) data without imputations. Explanatory variables are notated as originally found in the datasets. Clusters of data around SHAP value of zero indicates small impact on model output.

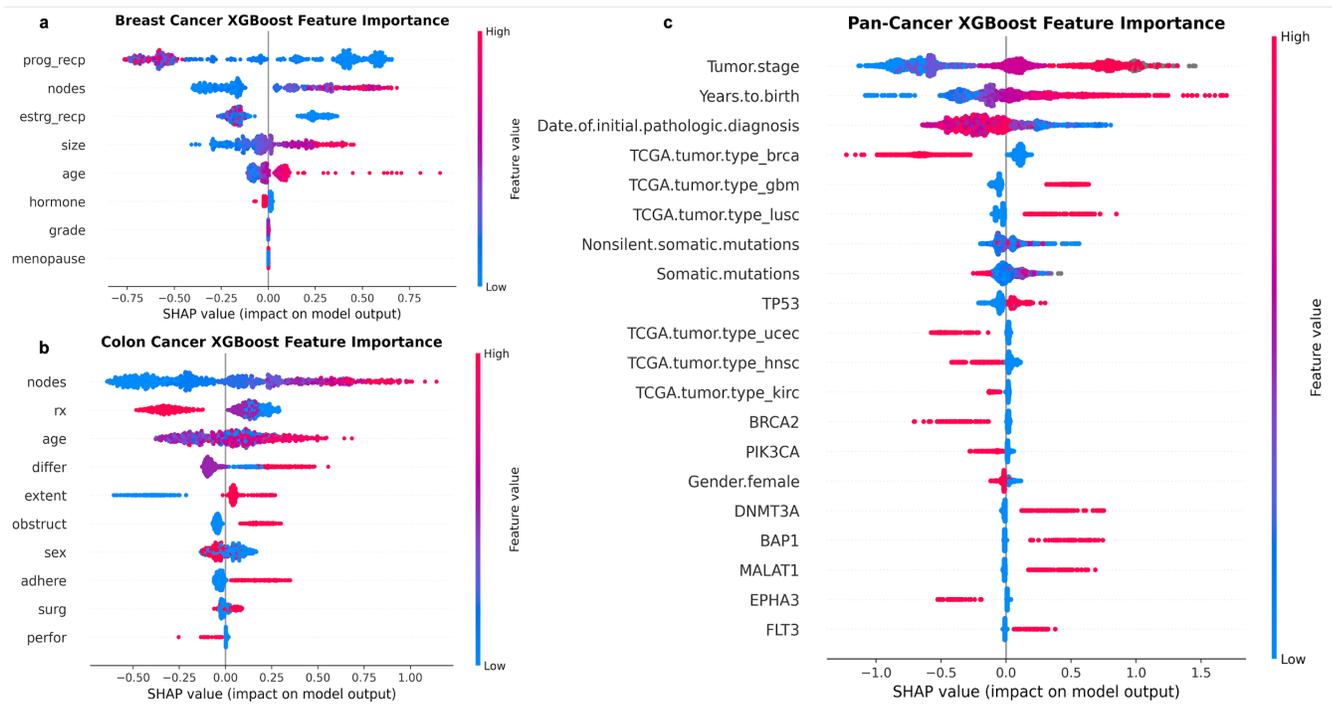

**Figure 4**: Comparison plots of CoxPH with imputations and XGBoost without imputations for all significant explanatory variables in each dataset. We denoted explanatory variables with truncated names so as to fit in the plots. Blue and orange stars represent significance on that variable for CoxPH and XGBoost respectively. The explanatory variables that have both blue and orange stars where both models agreed on their significance.

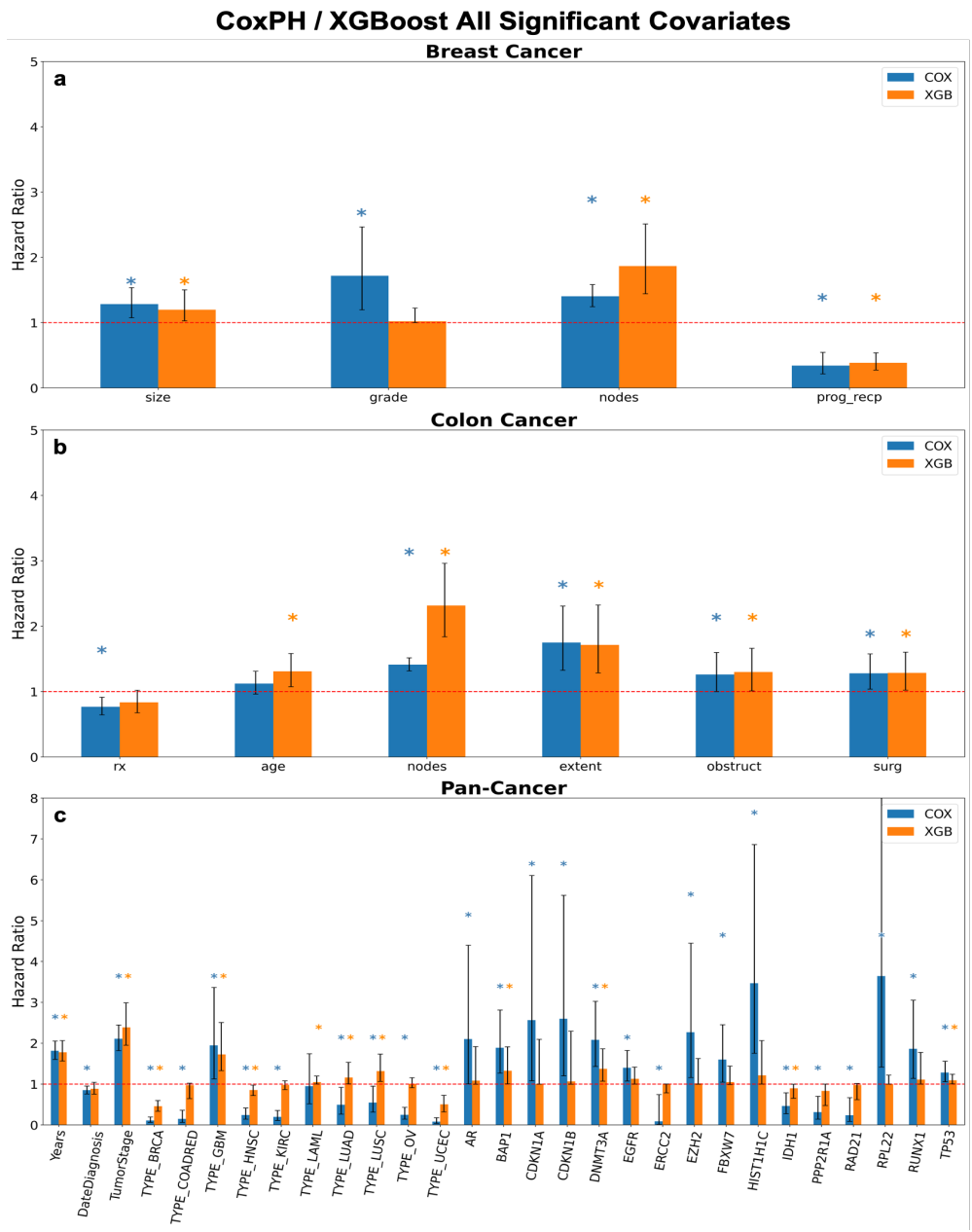

**Figure 5**: Kaplan-Meier estimates of survival for cancer types: lung adenocarcinoma (LUAD) and lung squamous cell carcinoma (LUSC) with their respective confidence intervals and the median survival are shown.

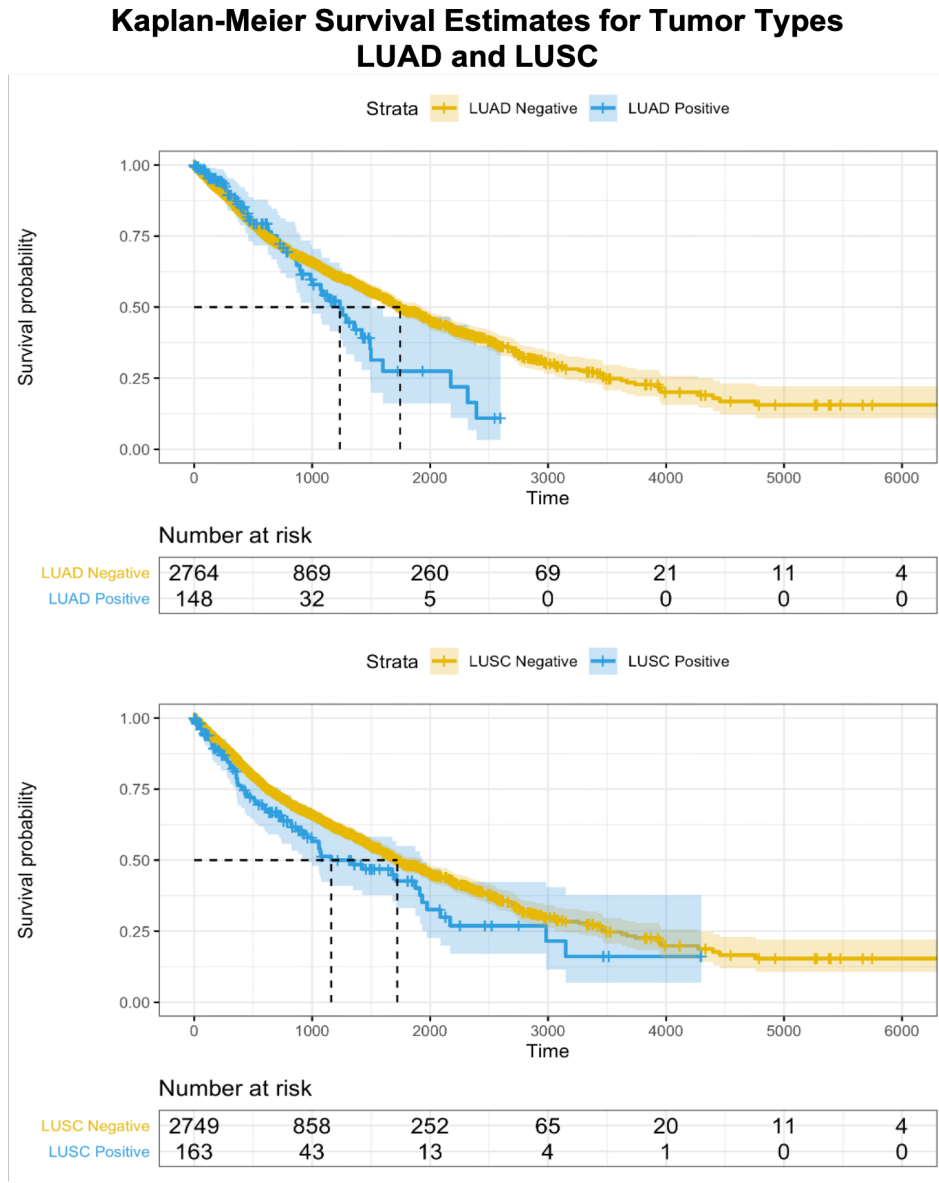